\documentclass[usenames,dvipsnames]{article}

\usepackage{nips15submit_e,times}
\usepackage{hyperref}
\usepackage{url}
\usepackage{subcaption}
\usepackage{graphicx}

\graphicspath{{figures/}}

\title{Can deep learning help you\\find the perfect match?}

\author{Harm de Vries\\
University of Montreal\\
\texttt{mail@harmdevries.com}
\And 
Jason Yosinski\\
Cornell University\\
\texttt{yosinski@cs.cornell.edu}
}

\nipsfinalcopy

\newif\iftodos

\todostrue

\iftodos
\newcommand{\todo}[1]{\textcolor{red}{[#1]}}
\newcommand{\done}[1]{\textcolor{Emerald}{[#1]}}
\newcommand{\comment}[1]{\textcolor{blue}{[#1]}}
\newcommand{\jby}[1]{\textcolor{Orange}{[JBY: #1]}}
\newcommand{\lowpriority}[1]{\textcolor{green}{[#1]}}
\else
\newcommand{\todo}[1]{}
\newcommand{\done}[1]{}
\newcommand{\comment}[1]{}
\newcommand{\jby}[1]{}
\newcommand{\lowpriority}[1]{}
\fi

\begin{document}
\maketitle
\begin{abstract}
  Is he/she my type or not? The answer to this question depends on the personal preferences of the one asking it. The individual process of obtaining a full answer may generally be difficult and time consuming, but often an approximate answer can be obtained simply by looking at a photo of the potential match. Such approximate answers based on visual cues can be produced in a fraction of a second, a phenomenon that has led to a series of recently successful dating apps in which users rate others positively or negatively using primarily a single photo.
  In this paper we explore using convolutional networks to create a model of an individual's personal preferences based on rated photos. This introduced task is difficult due to the large number of variations in profile pictures and the noise in attractiveness labels. Toward this task we collect a dataset comprised of $9364$ pictures and binary labels for each. We compare performance of convolutional models trained in three ways: first directly on the collected dataset, second with features transferred from a network trained to predict gender, and third with features transferred from a network trained on ImageNet.
  Our findings show that ImageNet features transfer best, producing a model that attains $68.1\%$ accuracy on the test set and is moderately successful at predicting matches.
\end{abstract}

\section{Introduction}
Online dating has become a popular way to seek partners. Big dating services, such as OKCupid.com and Match.com, have enormous numbers of users, so methods that can automatically filter users based on personal preferences can increase the probability of successful matches significantly. The aim of dating systems is to help you in this process by presenting the most promising profiles. The traditional way to recommend profiles is to calculate match scores that are based on social and physical attributes, e.g. body type and education level. A recently popular dating app, Tinder,\footnote{Available in 24 languages with an estimated user base of $50$ million.} employs an alternative matching strategy. Profile pictures\footnote{Also mutual interests and friends are shown, but most emphasis is put on pictures.} of geographically nearby users are presented one at a time, and a user can quickly decide to like or dislike the profile by swiping the screen right or left, respectively. If both users like each other, they are a match and have the ability to chat with each other to possibly arrange an offline date.

The success of these apps indicates the importance of visual appearance in the search for the ideal partner and highlights that matching algorithms based purely on self-reported attributes ignore important visual information. However, extracting visual information like attractiveness and personality type from a profile picture is a challenging task. Recently proposed matching algorithms \cite{Krzywicki201550,DBLP:journals/corr/abs-cs-0703042,akehurst2011ccr} sidestep this problem by using collaborative filtering. Instead of calculating matching scores based on the content of a profile, such systems recommend profiles that are high ranked by similar users. One of the drawbacks of collaborative filtering is that it suffers from the so-called cold start problem: when a new user enters the system it cannot make recommendations due to the lack of information. Understanding the content of the profile pictures could partially solve this cold start problem: we still can not recommend profiles to a new user, but we can recommend his/her profile to existing users.

The 2012 winning entry \cite{NIPS2012_4824}, often dubbed AlexNet, of the ImageNet competition \cite{Deng09imagenet:a} has rapidly changed the field of computer vision. Their convolutional network (convnet) trained on a large labeled image database significantly outperformed all classical computer vision techniques on a challenging object recognition task. The key ingredient of the success of convnets and other deep learning models is that they learn multiple layers of representations \cite{Bengio:2009:LDA:1658423.1658424} as opposed to hand-crafted or shallow features. Since 2012, several groups have improved upon the original convnet architecture \cite{DBLP:journals/corr/SimonyanZ14a,DBLP:journals/corr/SermanetEZMFL13,DBLP:journals/corr/SzegedyLJSRAEVR14} with the latest results achieving near-human level performance \cite{DBLP:journals/corr/HeZR015,DBLP:journals/corr/IoffeS15}.

Motivated by these recent advances, we investigate in this paper whether we can successfully train such deep learning models to predict personalized attractiveness scores from a profile picture. To this end, the author of this paper collected and labeled more than 9K profile pictures from dating app Tinder. We found, however, that the dataset was still too small to successfully train a convolutional network directly. We overcome this problem by using transfer learning, where we extract features using another neural network trained on a different task (for which more data is available). Several studies \cite{DBLP:journals/corr/DonahueJVHZTD13,DBLP:journals/corr/SimonyanZ14a,yosinski-2014-NIPS-how-transferable-are-features-in-deep} have demonstrated that high layer activations from top-performing ImageNet networks serve as excellent features for recognition tasks for which the network was not trained. The introduced attractiveness prediction task is defined over a very specific image distribution, namely profile pictures, possibly making transferability of ImageNet features rather limited. We therefore also investigate if transfer from another network -- one trained to predict gender from profile pictures -- is more effective.

\section{The task and data}
The aim of this project is to investigate whether we can predict preferences for potential partners solely from a profile picture. We take the first author as object of study. Although the results of one person can never be statistically significant, we consider it as a first step to study the feasibility of modern computer vision techniques to grasp a subtle concept such as attractiveness. 

\subsection{Attractiveness dataset}
In order to extract his preferences, the first author labeled $9364$ profile pictures from Tinder with binary labels: either like or dislike. Quite surprisingly, the dataset is fairly balanced with $53\%$ likes and $47\%$ dislikes. It seems unreasonable to be attracted to more than half of the population. We suspect that Tinder does not provide unbiased samples from the population but instead presents popular profiles more frequently.\footnote{There is clear incentive for Tinder to do so: the hope to match with more popular profiles keeps you using the application.} Another explanation is that mostly attractive people are using the application. Note that Tinder profiles contain up to six pictures, but that only the first one was viewed and labeled. The collected pictures were originally presented at a scale of $360 \times 360$, but they were later rescaled to $250 \times 250$ when training the convnet model for computational reasons.

During the process of labeling, the disadvantages of a binary labels became apparent.
Some profile pictures fell near the border of like and dislike, and in these cases
the ratings may have been affected by the mood of the subject.\footnote{We found that it also matters which profile pictures one have seen before; after a series of likes there is a tendency to keep liking.}
Unfortunately, this makes the attractiveness labeling quite noisy and thus harder to learn for any model. In order to quantify how much noise entered the labeling process, we performed another experiment a couple of weeks after the original labeling. This period was long enough to not remember or recognize the profile pictures. The first author classified $100$ random pictures from the dataset and compared them with the original labeling. He made $12$ errors out of $100$, achieving an ${\bf 88\%}$ accuracy on the original labeling. If we assume that these errors come from a $50/50$ guess on pictures near the classification boundary, we estimate that roughly a quarter ($12\cdot2/100 = .24$) of the profile pictures are not consistently labeled.

Another interesting question to ask is: how difficult is it for humans to learn the preference function of another? We investigate this question by setting up a small experiment with the second author of this paper, who trained on $100$ images and their corresponding labels. Training began by looking at all $50$ dislike and $50$ like pictures side by side, scrolling through them all a few times.
Then the training set was shuffled, pictures were displayed one a time, and the subject produced a label prediction after each photo.
The correct label was shown after each image, so that he could learn from his mistakes. This process was used to iterate through the training set four times, and accuracies over the four epochs were $86\%$, $82\%$, $88\%$, and $88\%$. Memorizing the last $12$ mistakes could definitely improve training performance, but this probably would not lead to better test set accuracy, so training was only carried out for four epochs. Test performance was then measured on the same $100$ random pictures as the above consistency experiment, with the subject making $24$ errors on the same images resulting in ${\bf 76\%}$ accuracy.
 
The results of this simple experiment gives a rough indication of the difficulty of the task, although we should be careful when interpreting these numbers. On the one hand, the preferences of the second author may be partially aligned with the first author, which could result in an overestimate of the ability for one human to learn the preferences of another. On the other hand, only $100$ pictures where given; perhaps with even more training images used, performance could increase further.

As a final note we stress that the collected profile pictures have much variation in viewpoints and personality types. In contrast to standard image recognition benchmarks, faces are not aligned and persons are not always in the center of the image. As we show in Section \ref{section:attractiveness}, this makes it difficult to train convnet directly on the small dataset. 

\begin{figure}[t]
\begin{subfigure}{0.3\textwidth}
\centering{
\includegraphics[width=3.5cm]{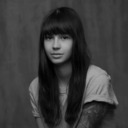}
\caption{Clean}}
\end{subfigure}
\begin{subfigure}{0.3\textwidth}
\centering{
\includegraphics[width=3.5cm]{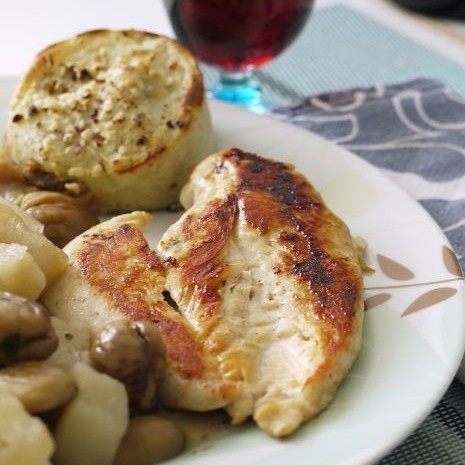}
\caption{Unknown}}
\end{subfigure}
\begin{subfigure}{0.3\textwidth}
\centering{
\includegraphics[width=3.5cm]{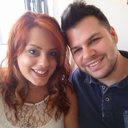}
\caption{Mixed}}
\end{subfigure}
\\
\begin{subfigure}{0.3\textwidth}
\centering{
\includegraphics[width=3.5cm]{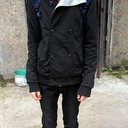}
\caption{No face}}
\end{subfigure}
\begin{subfigure}{0.3\textwidth}
\centering{
\includegraphics[width=3.5cm]{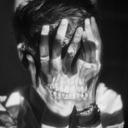}
\caption{Partial face}}
\end{subfigure}
\begin{subfigure}{0.3\textwidth}
\end{subfigure}
\caption{Example images of the categories encountered in our gender dataset. Note that above images are not from our dataset, but taken from \url{http://uifaces.com/} (a,c,d,e) and \url{http://www.morguefile.com/} (b) to illustrate the concepts.}
\label{figure:gender_categories}
\end{figure}

\subsection{Gender dataset}\label{section:gender_dataset}
As we describe in Section \ref{section:attractiveness}, we found that the collected attractiveness dataset is too small for a convolutional network to train on directly. This motivated the collection of a second dataset consisting of $418,452$ profile pictures from another dating site -- OKCupid -- where each user is labeled with a gender and age. To make training of this neural network straightforward, we created this dataset such that we have an equal number of male and female profile pictures. We discard age information in the following, because we found that the signal was too noisy.\footnote{Most OKCupid users fall in a relatively narrow range between 20 and 35, which makes it hard even for humans to accurately predict age. Also, users are not necessarily honest about their ages on OKCupid.} Our strategy is to train a convnet for gender prediction, and then transfer the learned feature representations to attractiveness prediction. 

The dataset was collected from a real-world dating site which raises questions about the quality of the provided labels. For example, some pictures might be wrongly labeled, or even impossible to discriminate for humans. It was too time consuming to clean up the full dataset, so we estimated the quality of the labels by randomly sampling $1000$ images from the gender dataset and categorizing them as one of the following:
\begin{description}
\item[Clean:] If the gender is clearly recognizable from the picture. 
\item[Unknown: ] If there isn't a person in the picture. Note that trained networks may still be able to infer gender from other objects in the picture (for example, if cars are more likely to occur in male vs. female profile pictures, the network may learn this).
\item[Mixed:] If both males and females appear in the picture. It's sometimes possible to infer the gender by looking at the leading person in the picture. 
\item[No face:] There is no face visible in the picture; only some body parts. For instance, if a picture is taken from far away and only the back is visible. 
\item[Partial face:] If most part of the face is not visible. For example, a close-up of the eye.
\end{description}
We provide examples of the categories in Figure \ref{figure:gender_categories}. The resulting numbers per category are given in Table \ref{table:categories}. We conclude that almost 90$\%$ of the pictures are clean, and the remaining $10\%$ are at least difficult. We may therefore guess that the maximum human performance at predicting gender from this specific dataset would be around $95\%$, with the last $5\%$ due to uninteresting factors. Moreover, our primary task is attractiveness prediction, thus learning the subtle uninteresting factors might not lead to better transferable features. 

As usual in prediction tasks, we randomly split the attractiveness and gender datasets into training, validation and test sets. For the attractiveness dataset we used $90\%$, $5\%$, and $5\%$ of the data for the training, validation and test set, respectively. Since we have more data for gender prediction, we make a $80\%$, $10\%$, and $10\%$ splits of that dataset.

\section{Experiments}
In Section \ref{section:attractiveness} we first train a convnet to predict attractiveness from the small labeled dataset. Section \ref{section:gender} presents the details of training a convnet for gender prediction. We then investigate in Section \ref{section:transfer} how well the features of this network transfer to attractiveness prediction. We compare against features obtained from VGGNet \cite{DBLP:journals/corr/SimonyanZ14a}, one of the top performing convnets on ImageNet. 

\begin{table}[t]
\caption{The resulting categories of inspection of $1000$ random samples from the dataset.}
\centering{
\begin{tabular}{p{3cm}|p{3cm}}
Category & Number\\
\hline
Clean & 895\\
Unknown & 25\\
Mixed & 44\\
No face & 25\\
Partial face & 11\\
\end{tabular}
}
\label{table:categories}
\end{table}

\subsection{Attractiveness prediction}\label{section:attractiveness}
After collecting the data, our first attempt was to train a convnet on the attractiveness dataset. Our architecture is inspired by VGGNet \cite{DBLP:journals/corr/SimonyanZ14a}, and follows the latest trends in architecture design to have very deep networks and small filter sizes. We use five convolutional layers, all with $3$x$3$ filter sizes and rectified linear activation functions. Each layer is followed with non-overlapping max pooling of size $2$x$2$. We start with $8$ feature maps in the first layer and gradually increase it to $32$ in the last convolutional layer. There are two fully connected layers on top of respectively $32$ and $16$ units. The network has on the order of $870K$ parameters. Architectural details are shown in Table \ref{table:architecture} (a).

The only preprocessing step that is applied is subtracting the training set mean from all images. We regularize the network by applying dropout \cite{hinton2012improving} with probability $0.5$ on the fully connected layers, and include $L2$ weight decay with coefficient $0.001$. The convnet is trained with Stochastic Gradient Descent (SGD) to minimize the negative log likelihood, optimization proceeds over $50$ epochs with a learning rate of $0.001$, $0.9$ momentum and a mini-batch size of $128$.  

Figure \ref{figure:error_curves1} (a) shows the training and validation misclassification rate during optimization. We can see that even this very small network with strong regularization immediately overfits. We think that there is simply too much variation in the profile pictures for the convnet to learn the regularities from the raw profile pictures. Hence, we decide not to explore further regularization techniques, but instead focus on transfer learning. In the next sections we investigate if a convnet trained for gender prediction results in good representations for attractiveness prediction. 

\begin{table}
\caption{The convnet architectures for (a) attractiveness prediction and (b) gender prediction. Conv-$m$x$m-n$ denotes a convolutional layer with filter size of $c$ by $c$ and $n$ feature maps. MaxPool-$m$x$m$ stands for a max pooling layer with non-overlapping $m$ by $m$ pool size, while FC-$n$ is an abbreviation of fully connected layer with $n$ outputs. }
\label{table:architecture}
\begin{subfigure}[t]{0.5\textwidth}
\vspace{0pt}
\centering{
\caption{Attractiveness prediction}
\begin{tabular}{|l|l|}
\hline
Input size & Layer\\
\hline\hline
$250$x$250$ &Conv$3$x$3-8$\\
$248$x$248$ &MaxPool-$2$x$2$\\
$124$x$124$ &Conv$3$x$3-16$\\
$122$x$122$ &MaxPool-$2$x$2$\\
$61$x$61$ &Conv$3$x$3-16$\\
$59$x$59$ &MaxPool-$2$x$2$\\
$30$x$30$ &Conv$3$x$3-32$\\
$28$x$28$ &MaxPool-$2$x$2$\\
$14$x$14$ &Conv$3$x$3-32$\\
$12$x$12$ &MaxPool-$2$x$2$\\
\hline
&FC-$32$\\
&FC-$16$\\
&FC-$2$\\
&Softmax\\
\hline
\end{tabular}}
\end{subfigure}\hfill
\begin{subfigure}[t]{0.5\textwidth}
\vspace{0pt}
\centering{
\caption{Gender prediction}
\begin{tabular}{|l|l|}
\hline
Input size & Layer\\
\hline\hline
$250$x$250$ &Conv$3$x$3-64$\\
$248$x$248$ &MaxPool-$2$x$2$\\
$124$x$124$ &Conv$3$x$3-128$\\
$122$x$122$&Conv$3$x$3-128$\\
$120$x$120$&MaxPool-$2$x$2$\\
$60$x$60$&Conv$3$x$3-256$\\
$58$x$58$&Conv$3$x$3-256$\\
$56$x$56$&MaxPool-$2$x$2$\\
$28$x$28$&Conv$3$x$3-512$\\
$26$x$26$&Conv$3$x$3-512$\\
$24$x$24$&MaxPool-$2$x$2$\\
$12$x$12$&Conv$3$x$3-512$\\
$10$x$10$&Conv$3$x$3-512$\\
\hline
&FC-$1024$\\
&FC-$512$\\
&FC-$2$\\
&Softmax\\
\hline
\end{tabular}}
\end{subfigure}
\end{table}

\subsection{Gender prediction}\label{section:gender}
The gender dataset with over 400k images is much bigger than the attractiveness dataset. Therefore, we can afford to train a much bigger network without the risk of over fitting. The proposed convnet architecture is similar in spirit to the attractiveness network presented in the previous section. We decide to use nine convolutional layers with $3$x$3$ filter sizes and rectified linear activation functions. We further apply $2$x$2$ max pooling after two convolutional layer, except for the first layer where we directly apply pooling after one layer. We follow the rule of thumb introduced in \cite{DBLP:journals/corr/SimonyanZ14a} and double the number of feature maps after each pooling layer, except for the last pooling layer where we kept the number of feature maps the same. The biases (in contrast to the weights) in the convolutional layers are untied i.e. each location in a feature map has its own bias parameter. The final $12$-layer architecture is shown in Table \ref{table:architecture} (b), and has over $28$ million parameters. 

We tried several small modifications on this architecture: decreasing the number of feature maps (starting from $32$), using tied biases, and adding an extra pooling after the two final convolutional layers. However, we obtained the best performance with the network described above. 

\begin{figure}[t]
\begin{subfigure}{0.5\textwidth}
\includegraphics[width=7.5cm]{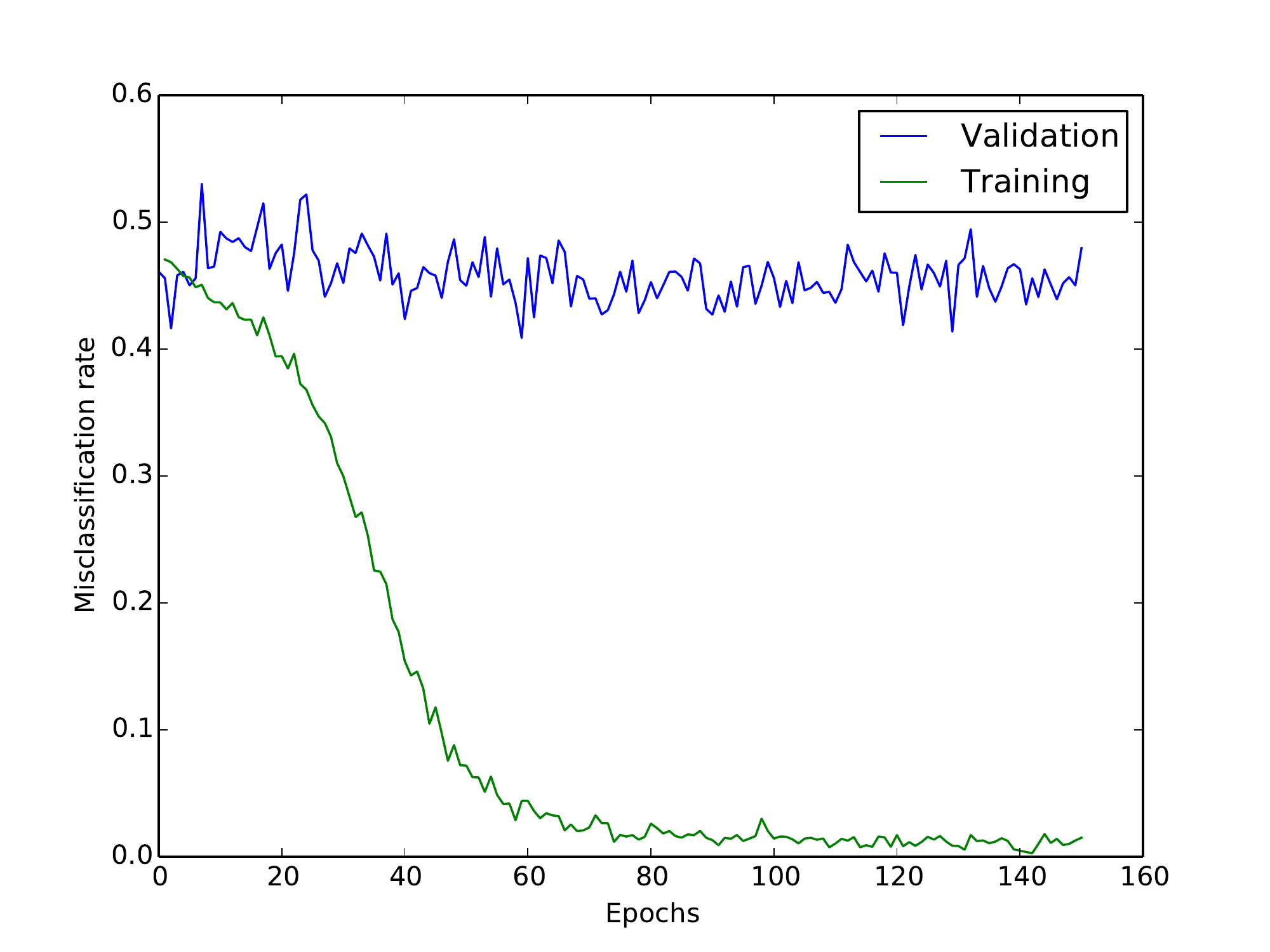}
\caption{Attractiveness}
\end{subfigure}
\begin{subfigure}{0.5\textwidth}
\includegraphics[width=7.5cm]{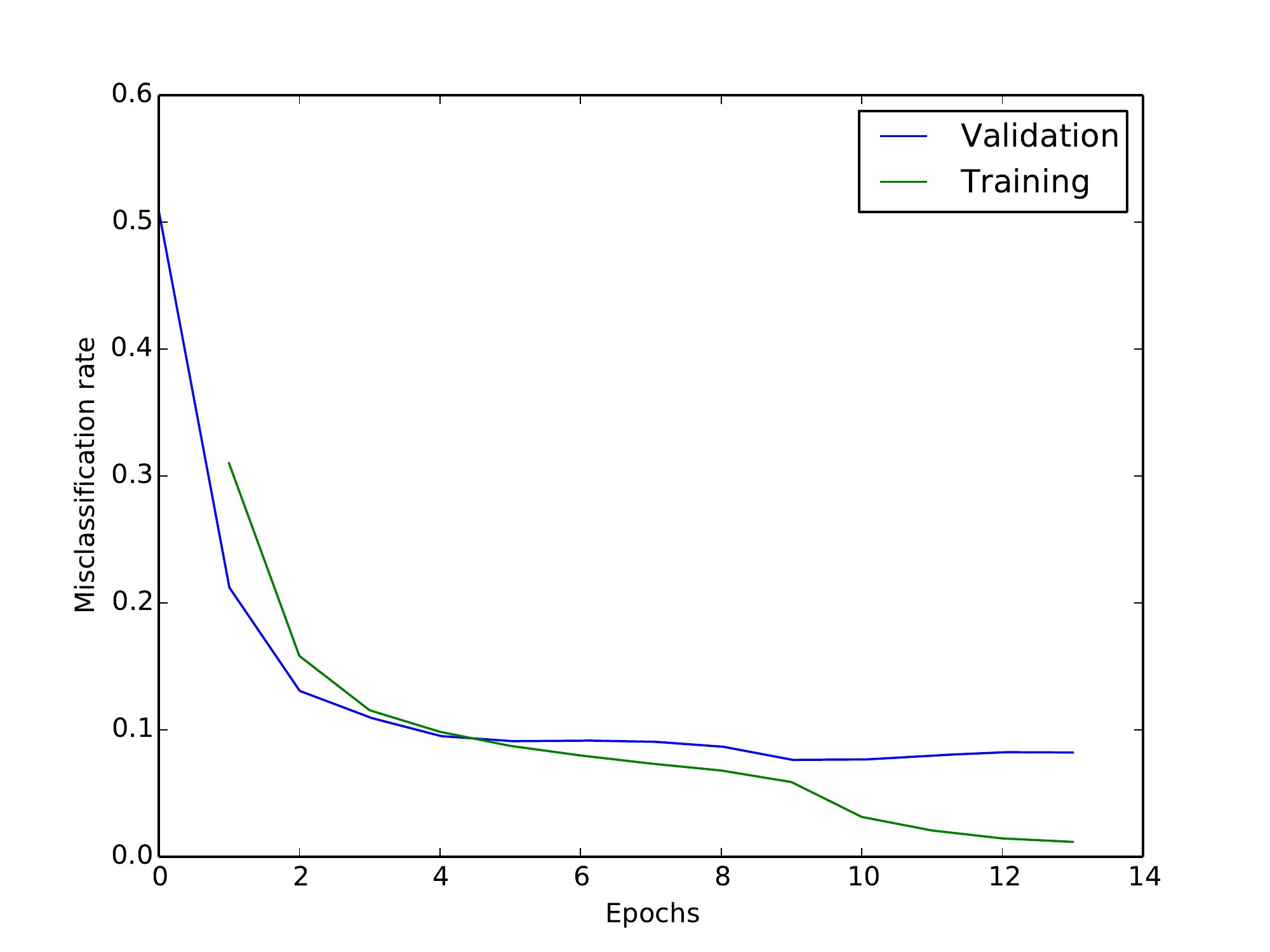}
\caption{Gender}
\end{subfigure}
\caption{The training and validation error curves for a) attractiveness  prediction and b) gender prediction. }
\label{figure:error_curves1}
\end{figure}

We also apply dropout with probability $0.5$ on the fully connected layers, and include $L2$ weight decay with coefficient $0.0001$. The weights are initialized from $\mathcal{U}(-0.06, 0.06)$, while the biases are initially set to zero.  We again train with Stochastic Gradient Descent (SGD) to minimize the negative log likelihood. We optimized for $13$ epochs with a learning rate of $0.001$, $0.9$ momentum, and a mini-batch size of $50$. The models were implemented in Theano \cite{Bastien-Theano-2012} and took about $3$ days to train on a GeForce GTX Titan Black. The misclassification rates during training are shown in Figure \ref{figure:error_curves1} (b). Note that in this figure the training error is aggregated over mini-batches, and only gives us a rough estimate of the true training error.  

The final model was selected by early stopping at epoch $9$, and achieved $7.4\%$ and $7.5\%$ error on the validation and test set, respectively. In Section \ref{section:gender_dataset} we established that approximately $10\%$ of the dataset is difficult. Hence, we consider $92.5\%$ accuracy as very good performance, likely approaching that which would be obtained by a human.

\subsection{Transfer learning}\label{section:transfer}
We compare two transfer learning strategies: one from the gender net and the other from VGGNet, one of the top-performing ImageNet convnets.

\subsubsection{Gender}
After training the gender network we explore if the features are helpful to predict attractiveness. The gender network has approximately $28$ million parameters, and the available attractiveness dataset is relatively small, so training the full network probably leads to overfitting. We therefore decide to train only the last layers of the gender network. We compare training the last, the last two, and the last three layers, which have 1026, 525k, and 8.9m parameters, respectively. We do not apply dropout when training these last layers, but we do use the same $L2$ regularization as in the gender network. We train with SGD for $50$ epochs with a learning rate of $0.001$, $0.9$ momentum and a batch-size of $16$.

The training and validation curves are shown in Figure \ref{figure:error_curves2} (a-c). Note that the transfer performance is rather poor. Training only the last layer barely decreases the training error and significantly underfits. On the other hand, training all fully connected layers does decrease the training error very quickly, but doesn't carry over to the validation error. With early stopping on the validation error, we achieved the best performance of $61.3\%$ accuracy on the test set by only training the last two layers.

\subsubsection{ImageNet}
\begin{figure}[t]
\begin{subfigure}{0.5\textwidth}
\centering{
\includegraphics[width=7.5cm]{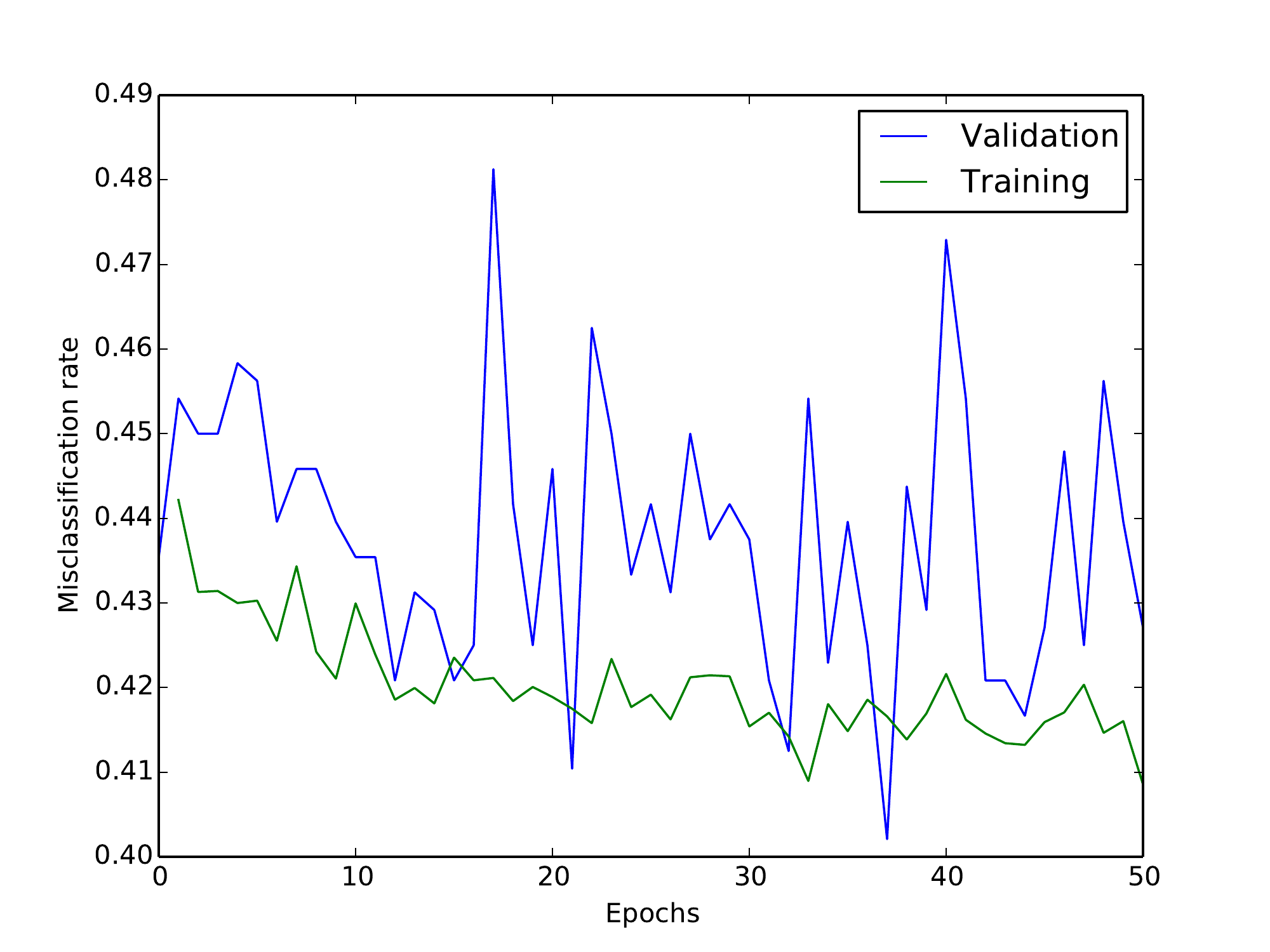}
\caption{Transfer from gender, fine-tune last layer}}
\end{subfigure}
\begin{subfigure}{0.5\textwidth}
\centering{
\includegraphics[width=7.5cm]{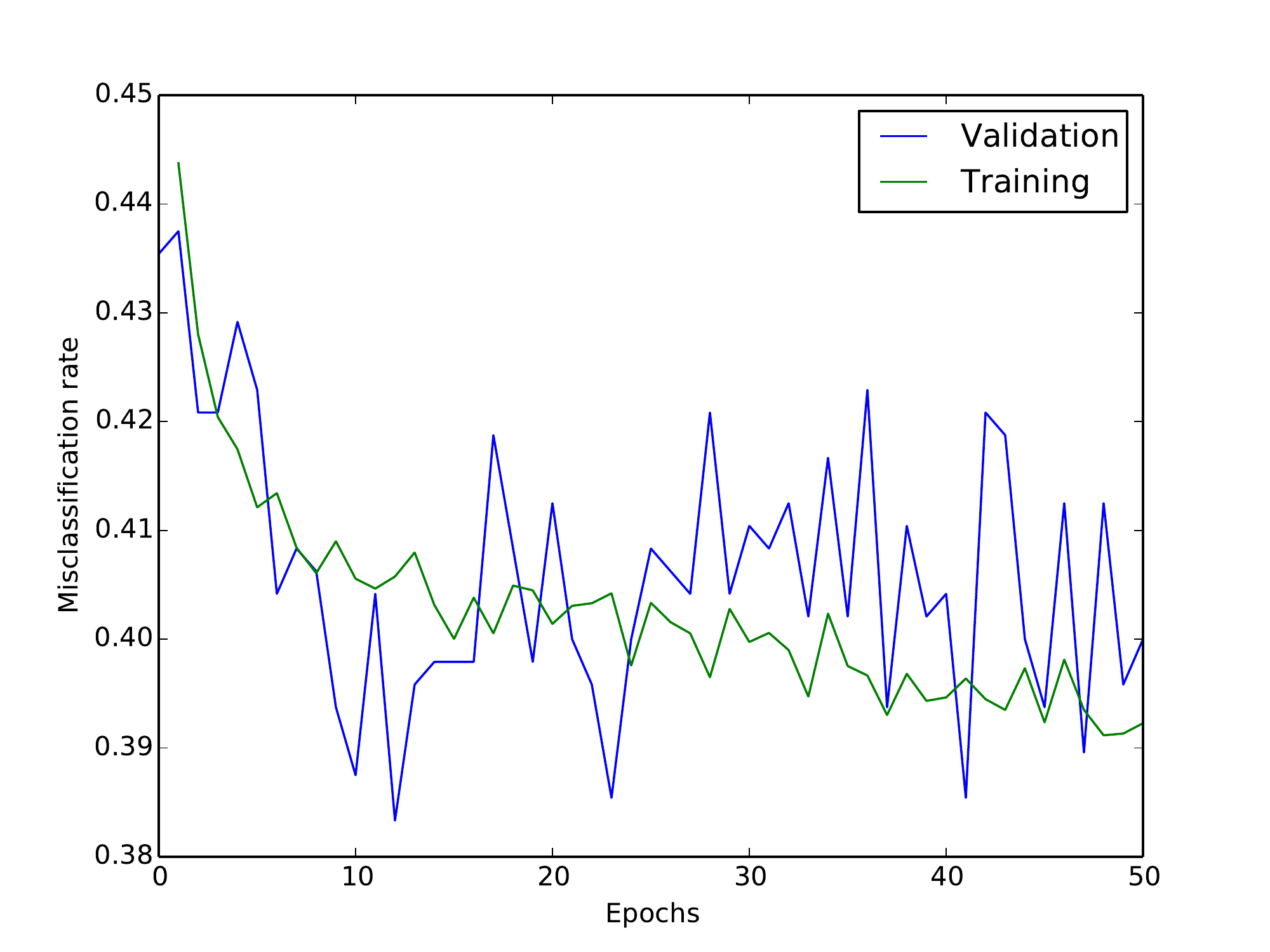}
\caption{Transfer from gender, fine-tune last two layers}}
\end{subfigure}
\begin{subfigure}{0.5\textwidth}
\centering{
\includegraphics[width=7.5cm]{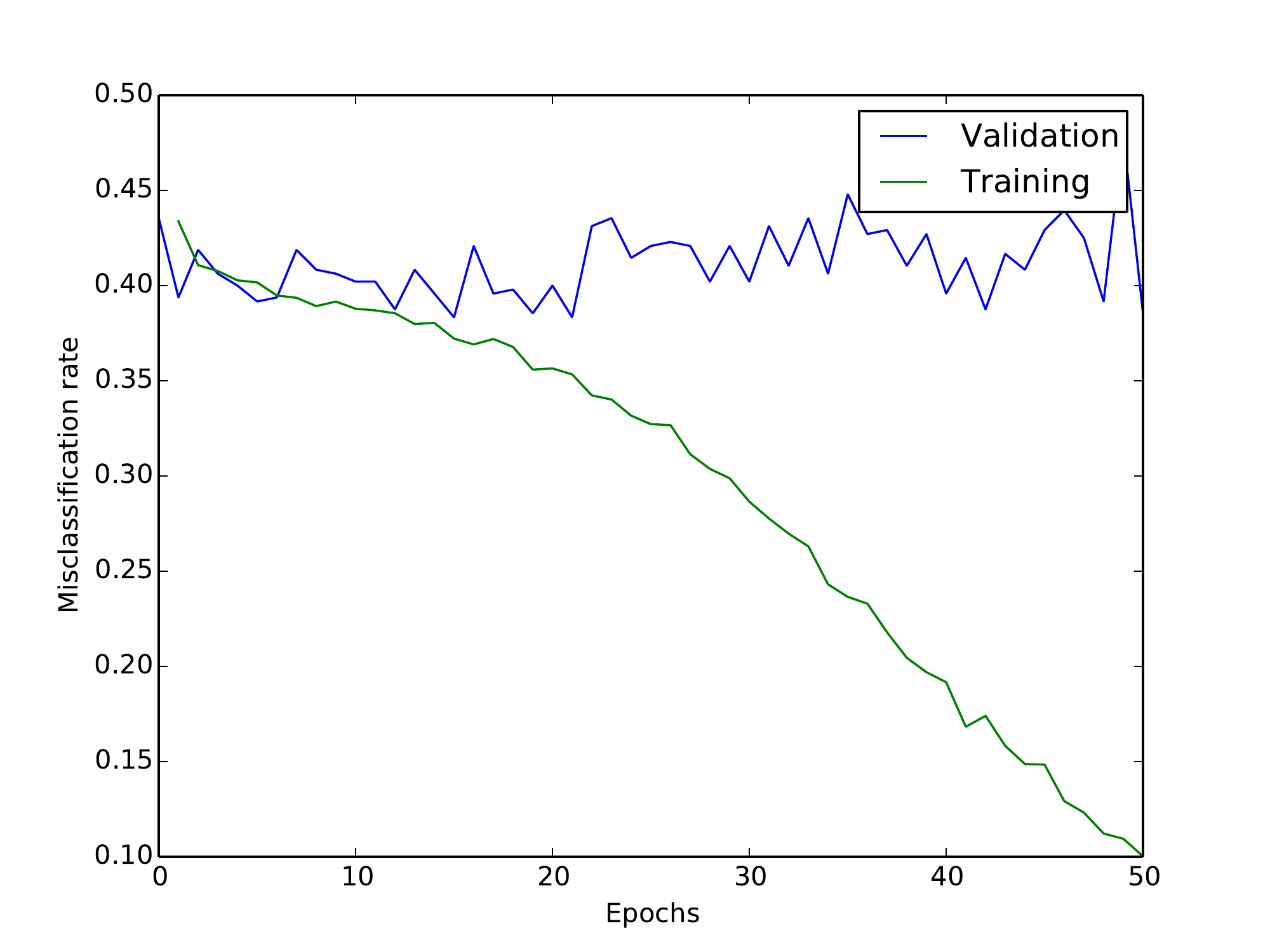}
\caption{Transfer from gender, fine-tune last three layers}} 
\end{subfigure}
\begin{subfigure}{0.5\textwidth}
\centering{
\includegraphics[width=7.5cm]{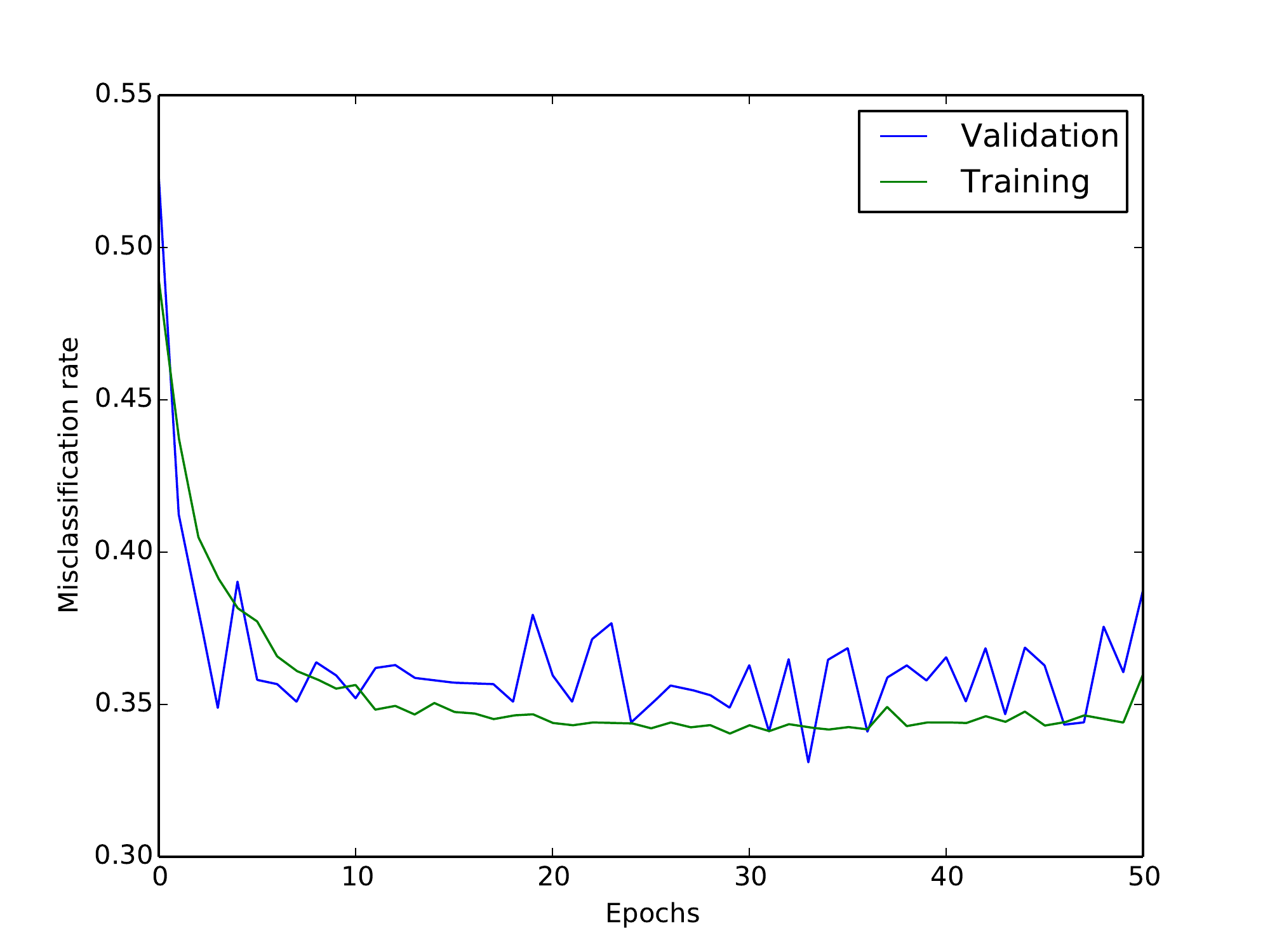}
\caption{Transfer from VGGNet}}
\end{subfigure}
\caption{The training and validation error curves for attractiveness prediction by training a) the last layer from our gender network and b) the last two layers c) the last three layers and d) a logistic regression on top of VGGNet features. }
\label{figure:error_curves2}
\end{figure}

The features extracted from ImageNet networks are known to achieve excellent transfer performance \cite{DBLP:journals/corr/DonahueJVHZTD13,yosinski-2014-NIPS-how-transferable-are-features-in-deep}. We decide to use VGGNet \cite{DBLP:journals/corr/SimonyanZ14a}, one of the top performing ImageNet convnets, and use Caffe \cite{jia2014caffe} to extract the features. In order to feed the images to VGGNet, we resize all images to $224$x$224$. We extract $4096$ dimensional features from the highest layer (called FC7) of the $19$-layer VGGNet. We put a logistic regression with weight decay on top of the extracted representation. After finetuning the hyperparameters, we obtained the best results with a $L2$ regularization coefficient of $0.8$, a learning rate of $0.0001$, and momentum of $0.9$. Note that the relatively strong weight decay is used to prevent overfitting. The error curves during training are presented in Figure \ref{figure:error_curves2} (d). We again apply early stopping on the validation error. Our best model obtains an validation and test accuracy of $66.9\%$ and $68.1\%$, respectively.

\section{Discussion and Conclusion}
The VGGNet features clearly outperform the features obtained from the gender prediction task. Our findings confirm that ImageNet activations are excellent image features for a wide variety tasks. However, we did not expect them to outperform the features from the gender prediction task since that network was trained on a similar set of images. One possible explanation for the poor transfer is that the gender network learns features that are invariant to within-class (female or male) characteristics and are therefore not appropriate to discriminate between within-class profile pictures (here: female). Another reason could be that the gender network only has two classes, which does not force the network to learn very rich features. Possible directions for future research are to investigate if adding extra classes of non-profile pictures or other labels to the profile pictures would lead to better transferable features.

Further studies could also investigate other ways to deal with the huge variability in the profile pictures. For example, face extraction could be a good way to reduce variability, while keeping the most important aspect of attractiveness. We believe that the most promising avenue is to collect a bigger and cleaner dataset from which a better feature representation for attractiveness prediction could be learned. It remains an open question what kind of label information could lead to the best learned representation for predicting attractiveness. For now though, even pretraining on the semantically distant cats, dogs, automobiles, etc. of ImageNet provides features rich enough to predict at 68\% accuracy, which covers about half of the gap between a random prediction (50\%) and human labels (88\%).

\section{Acknowledgement}
We thank Mehdi Mirza for extracting the VGGNet features. We also thank the developers of Theano \cite{Bastien-Theano-2012} and Blocks \cite{blocksfuel}, the computational resources provided by Compute Canada and Calcul Qu\'{e}bec, and the NASA Space Technology Research Fellowship (JY) for funding.
We are grateful to many members of and visitors to the LISA lab for helpful discussions, in particular to Yoshua Bengio, Aaron Courville, Roland Memisevic, Kyung Hyun Cho, Yann Dauphin, Laurent Dinh, Kyle Kastner, Junyoung Chung, Julian Serban, Alexandre de Br\'{e}bisson, C\'{e}sar Laurent, and Christopher Olah. 

\bibliographystyle{abbrv}
\bibliography{refs}

\end{document}